\RequirePackage{fix-cm}
\documentclass[smallextended]{svjour3}       
\smartqed  

\usepackage[noadjust]{cite}

\usepackage{ulem}
\usepackage{color}
\usepackage{graphicx}
\usepackage{subfigure}
\usepackage{bbding}
\usepackage[colorlinks,
           linkcolor=black,
           anchorcolor=blue,
           citecolor=blue
           ]{hyperref} 
\begin{document}

\title{Deep Learning to Segment Pelvic Bones: Large-scale CT Datasets and Baseline Models
}


\author{Pengbo Liu \and
	        Hu Han \and
        Yuanqi Du \and
        Heqin Zhu \and
        Yinhao Li \and
        Feng Gu \and
        Honghu Xiao \and
        Jun Li \and
        Chunpeng Zhao \and
        Li Xiao \and
        Xinbao Wu \and
        S. Kevin Zhou 
}


\institute{Pengbo Liu, Hu Han, Heqin Zhu, Yinhao Li, Feng Gu, Jun Li, Li Xiao, S.Kevin Zhou \at
              Institute of Computing Technology, Chinese Academy of Sciences, Beijing, China\\
              \email{s.kevin.zhou@gmail.com}
          \and
          Honghu Xiao, Chunpeng Zhao, Xinbao Wu \at
              Beijing Jishuitan Hospital, Beijing, China
            \and
            Yuanqi Du \at
             George Mason University, Virginia, USA
             \and
             Feng Gu \at
             Beijing Electronic Science and Technology Institute, Beijing, China
}

\date{Received: date / Accepted: date}

\maketitle

\begin{abstract}

\textit{Purpose:} Pelvic bone segmentation in CT has always been an essential step in clinical diagnosis and surgery planning of pelvic bone diseases. Existing methods for pelvic bone segmentation are either  hand-crafted or semi-automatic and achieve limited accuracy when dealing with image appearance variations due to the multi-site domain shift, the presence of contrasted vessels, coprolith and chyme, bone fractures, low dose, metal artifacts, etc. Due to the lack of a large-scale pelvic CT dataset with annotations, deep learning methods are not fully explored. 

\textit{Methods:} In this paper, we aim to bridge the data gap by curating a large pelvic CT dataset pooled from multiple sources, including $1,184$ CT volumes with a variety of appearance variations. Then we propose for the first time, to the best of our knowledge, to learn a deep multi-class network for segmenting lumbar spine, sacrum, left hip, and right hip, from multiple-domain images simultaneously to obtain more effective and robust feature representations. 
Finally, we introduce a post-processor based on the signed distance function (SDF).

\textit{Results:} Extensive experiments on our dataset demonstrate the effectiveness of our automatic method, achieving an average Dice of 0.987 for a metal-free volume. SDF post-processor yields a decrease of 15.1\% in Hausdorff distance compared with traditional post-processor.

\textit{Conclusion:} We believe this large-scale dataset will promote the development of the whole community and open source the images, annotations, codes, and trained baseline models at \url{https://github.com/ICT-MIRACLE-lab/CTPelvic1K}\label{opensource}.

\end{abstract}

\section{Introduction}
\label{intro}
The pelvis is an important structure connecting the spine and lower limbs and plays a vital role in maintaining the stability of the body and protecting the internal organs of the abdomen. The abnormality of the pelvis, like hip dysplasia~\cite{HipDysplasia} and pelvic fractures~\cite{2018pelvicFracture}, can have a serious impact on our physical health. For example, as the most severe and life-threatening bone injuries, pelvic fractures can wound other organs at the fracture site, and the mortality rate can reach 45\%~\cite{2020openpelvic} at the most severe situation, the open pelvic fractures. Medical imaging~\cite{zhou2021review,zhou2019handbook} plays an important role in the whole process of diagnosis and treatment of patients with pelvic injuries. Compared with X-Ray images, CT preserves the actual anatomic structure including depth information, providing more details about the damaged site to surgeons, so it is often used for 3D reconstruction to make follow-up \textbf{surgery planning} and evaluation of postoperative effects. In these applications, \textbf{accurate pelvic bone segmentation} is crucial for assessing the severity of pelvic injuries and helping surgeons to make correct judgments and choose the appropriate surgical approaches. In the past, surgeons segmented pelvis manually from CT using software like Mimics\footnote{\url{https://en.wikipedia.org/wiki/Mimics}}, which is time-consuming and non-reproducible. To address these clinical needs, we here present an automatic algorithm that can accurately and quickly segment pelvic bones from CT.

\begin{figure}[t]
\centering
\includegraphics[width=0.8\textwidth]{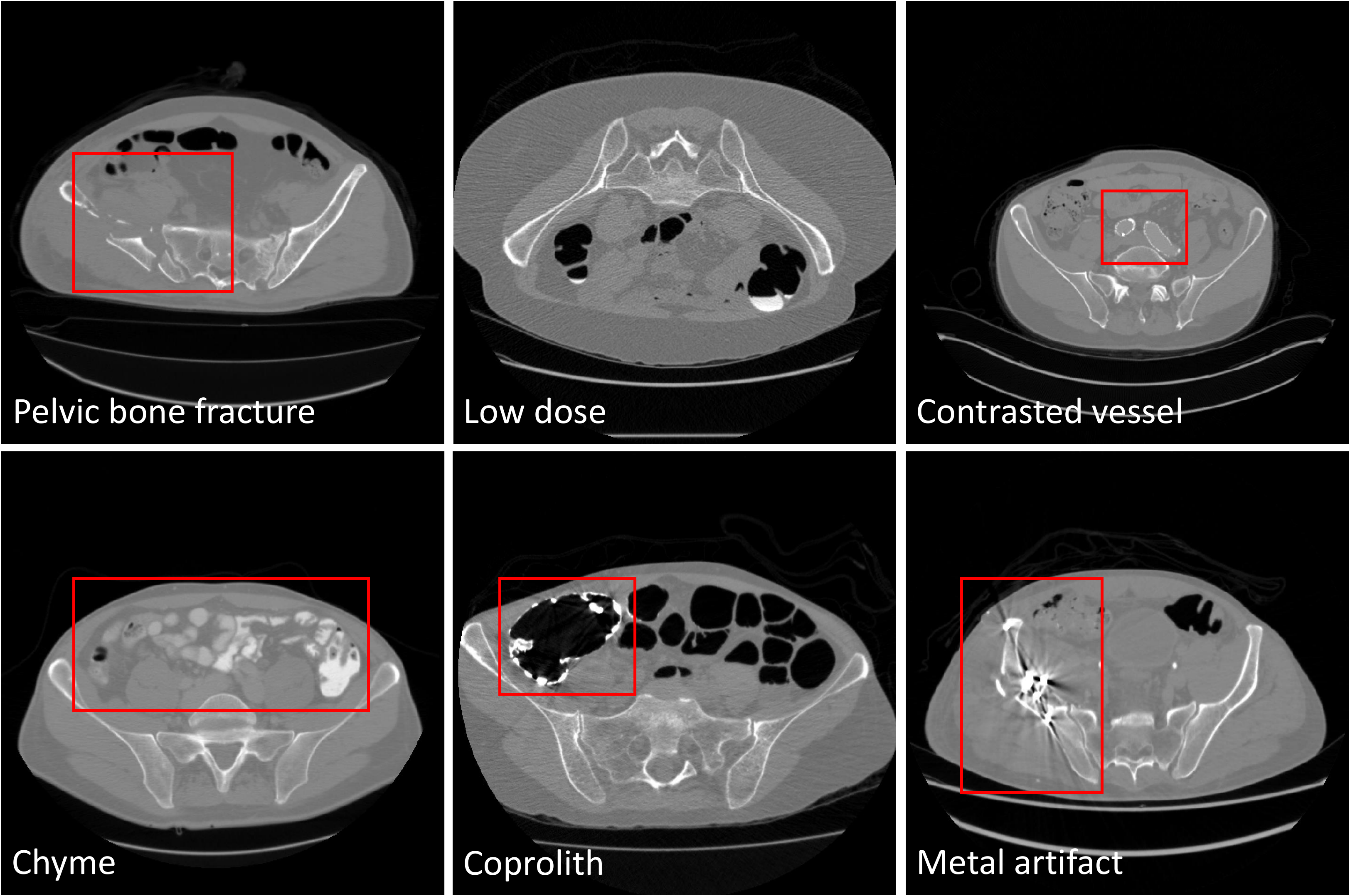}
\caption{Pelvic CT image examples with various conditions.} \label{fig-pic-3}
\end{figure}

Existing methods for pelvic bone segmentation from CT mostly use simple thresholding~\cite{gaussianThreshold_a6}, region growing~\cite{wavelet_a7}, and handcrafted models, which include deformable models~\cite{deformable_a8,activeContour_a9}, statistical shape models~\cite{shapeModel_a10,3DshapeModel_sta1}, watershed~\cite{keyframes_a11} and others~\cite{mriPelvicStruc_tra1,knowledgeOrganSpecificStrategies_tra2,femurSeg_tra3,femurXray_tra4,RFregressionVoting_rfr1,RF&HsparseShape_rfr2}. These methods focus on local gray information and have limited accuracy due to the density differences between cortical and trabecular bones. And trabecular bone is similar to that of the surrounding tissues in terms of texture and intensity. Bone fractures, if present, further lead to weak edges. Recently, deep learning-based methods~\cite{FCN_a13,unet_a12,FabianNNUnet_a4,zhao2017pyramid_dl1,atrousSeparableConvolution_dl2,3dUNet_dl3,pancreaticCyst_dl4,DenseVNet_dl5} have achieved great success in image segmentation; however, their effectiveness for CT pelvic bone segmentation is not fully known. Although there are some datasets related to pelvic bone~\cite{lee2012virtual_pelvicBoneDataset1,wu2016segmentation_pelvicBoneDataset2,hemke2020deep_pelvicBoneDataset3,chandar2016segmentation_pelvicBoneDataset4}, only a few of them are open-sourced and with small size (less than 5 images or 200 slices), far less than other organs~\cite{kits19_url3,MSD}. Although ~\cite{hemke2020deep_pelvicBoneDataset3} conducted experiments based on deep learning, the result was not very good (Dice=0.92) with the dataset only having 200 CT slices. For the robustness of the deep learning method, it is essential to have a comprehensive dataset that includes as many real scenes as possible. In this paper, we bridge this gap by curating a large-scale CT dataset and explore the use of deep learning in this task, which marks, to the best of our knowledge, \textbf{the first real attempt} in this area, with more statistical significance and reference value.

To build a comprehensive dataset, we have to deal with diverse image appearance variations due to differences in imaging resolution and field-of-view (FOV), domain shift arising from different sites, the presence of contrasted vessels, coprolith and chyme, bone fractures, low dose, metal artifacts, etc.  Fig. \ref{fig-pic-3} gives some examples about these various conditions. Among the above-mentioned appearance variations, the challenge of the metal artifacts is the most difficult to handle. Further, we aim at a multi-class segmentation problem that separates the pelvis into multiple bones, including \textit{lumbar spine}, \textit{sacrum}, \textit{left hip}, and \textit{right hip}, instead of simply segmenting out the whole pelvis from CT.

The contributions of this paper are summarized as follows:
\begin{itemize}
\item A \textit{pelvic CT dataset} pooled from multiple domains and different manufacturers, including $1,184$ CT volumes (over 320K CT slices) of diverse appearance variations (including 75 CTs with metal artifacts). Their multi-bone labels are carefully annotated by experts. We open source it to benefit the whole community;
\item Learning a \textit{deep multi-class segmentation network}~\cite{FabianNNUnet_a4} to obtain more effective representations for joint lumbar spine, sacrum, left hip, and right hip segmentation from multi-domain labeled images, thereby yielding desired accuracy and robustness;
\item A \textit{fully automatic analysis pipeline} that achieves high accuracy, efficiency, and robustness, thereby enabling its potential use in clinical practices.
\end{itemize}

\section{Our Dataset}
\label{sec:1}

\begin{table}[t]
\centering
\caption{Overview of our large-scale Pelvic CT dataset. `\#' represents the number of 3D volumes. `Tr/Val/Ts' denotes training/validation/testing set. Ticks[\Checkmark] in table refer to we can access the CT images' acquisition equipment manufacturer[M] information of that sub-dataset. Due to the difficulty of labeling the CLINIC-metal, CLINIC-metal is taken off in our supervised training phase.}\label{tab0}
\newsavebox{\tablebox}
\begin{lrbox}{\tablebox}
\begin{tabular}{lrcccc}
\hline
Dataset name[M] &  \#     & Mean spacing(mm) & Mean size  & \# of Tr/Val/Ts & Source and Year \\
\hline
ABDOMEN &  35 & (0.76, 0.76, 3.80) & (512, 512, 73)  & 21/7/7 & Public 2015\\
\hline
COLONOG[\Checkmark] &  731 &  (0.75, 0.75, 0.81) & (512, 512, 323) & 440/146/145& Public 2008\\
\hline
MSD\_T10 &  155 & (0.77, 0.77, 4.55) & (512, 512, 63)  & 93/31/31&  Public 2019\\
\hline
KITS19 &  44 &  (0.82, 0.82, 1.25) & (512, 512, 240) & 26/9/9&  Public 2019\\
\hline
CERVIX &  41 &  (1.02, 1.02, 2.50) & (512, 512, 102) &24/8/9 & Public 2015 \\
\hline 
CLINIC[\Checkmark] &  103 &  (0.85, 0.85, 0.80) & (512, 512, 345) & 61/21/21& Collected 2020\\
\hline
CLINIC-metal[\Checkmark] & 75 & (0.83, 0.83, 0.80) &(512, 512, 334) & 0(61)/0/14 & Collected 2020\\
\hline\hline
Our Datasets & $1,184$ &(0.78, 0.78, 1.46) &(512, 512, 273) & 665(61)/222/236 & -\\
\hline
\end{tabular}
\end{lrbox}
\scalebox{0.8}[0.8]{\usebox{\tablebox}}
\end{table}

\subsection{Data Collection}

To build a comprehensive pelvic CT dataset that can replicate practical appearance variations, we curate a large dataset of pelvic CT images from seven sources, two of which come from a clinic and five from existing CT datasets~\cite{COLONOGRAPHY,kits19_url3,MSD,matlas}. The overview of our large dataset is shown in Table~\ref{tab0}. 
These seven sub-datasets are curated separately from different sites and sources with different characteristics often encountered in the clinic. In these sources, we exclude some cases of very low quality or without pelvic region and remove the unrelated areas outside the pelvis in our current dataset. 

Among them, the raw data of COLONOG, CLINIC, and CLINIC-metal are stored in a DICOM format, with more information like scanner manufacturers can be accessed. More details about our dataset are given in Online Resource 1\footnote{\url{https://drive.google.com/file/d/115kLXfdSHS9eWxQmxhMmZJBRiSfI8_4_/view}\label{em1}}.

We reformat all DICOM images to NIfTI to simplify data processing and de-identify images, meeting the institutional review board (IRB) policies of contributing sites. All existing sub-datasets are under Creative Commons license CC-BY-NC-SA at least and we will keep the license unchanged. For CLINIC and CLINIC-metal sub-datasets, we will open-source them under Creative Commons license CC-BY-NC-SA
4.0.

\begin{figure}[t]
	\centering
	\includegraphics[width=0.9\columnwidth]{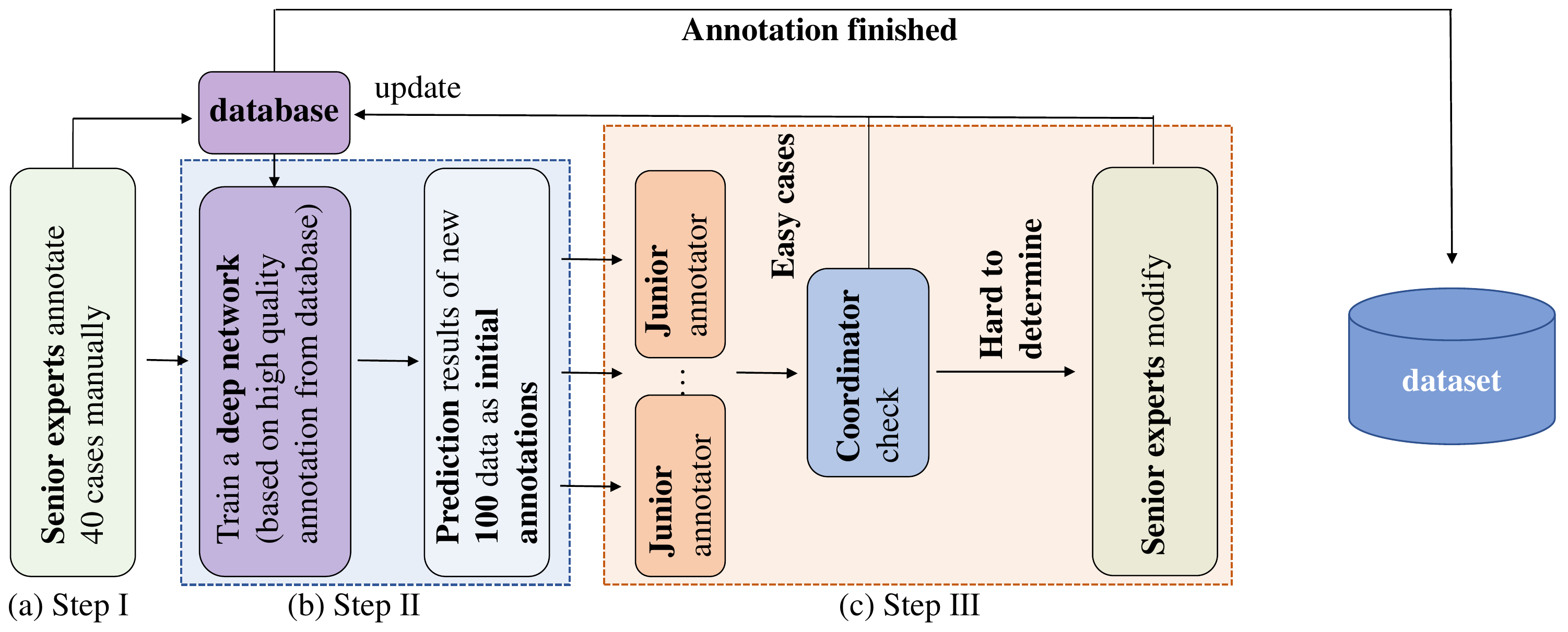}
	\caption{The designed annotation pipeline based on an AID (Annotation by Iterative Deep Learning) strategy. In Step \uppercase\expandafter{\romannumeral1}, two senior experts first manually annotate 40 cases of data as the initial database. In Step \uppercase\expandafter{\romannumeral2}, we train a deep network based on the human annotated database and use it to predict new data. In Step \uppercase\expandafter{\romannumeral3}, initial annotations from the deep network are checked and modified by human annotators. Step \uppercase\expandafter{\romannumeral2} and Step \uppercase\expandafter{\romannumeral3} are repeated iteratively to refine a deep network to a more and more powerful `annotator'. This deep network `annotator' also unifies the annotation standards of different human annotators.} \label{fig-pic-10}
\end{figure}

\subsection{Data Annotation}
Considering the scale of thousands of cases in our dataset and annotation itself is truly a subjective and time-consuming task. We introduce a strategy of Annotation by Iterative Deep Learning
(AID)~\cite{AIL} to speed up our annotation process. In the AID workflow, we train a deep network with a few precisely annotated data in the beginning. Then the deep network is used to automatically annotate more data, followed by revision from human experts. The human-corrected annotations and their corresponding images are added to the training set to retrain a more powerful deep network. These steps are repeated iteratively until we finish our annotation task. In the last, only minimal modification is needed by human experts. 

The annotation pipeline is shown in Fig.~\ref{fig-pic-10}. In Step \uppercase\expandafter{\romannumeral1}, two senior experts are invited to pixel-wise annotate 40 cases of CLINIC sub-dataset precisely as the initial database based on the results from simple thresholding method, using ITK Snap (Philadelphia, PA) software. All annotations are performed in the transverse plane. The sagittal and coronal planes are used to assist the judgment in the transverse plane. The reason for starting from the CLINIC sub-dataset is that the cancerous bone and surrounding tissues exhibit similar appearances at the fracture site, which needs more prior knowledge guidance from doctors. In Step \uppercase\expandafter{\romannumeral2}, we train a deep network with the updated database and make predictions on new 100 data selected randomly at a time. In Step \uppercase\expandafter{\romannumeral3}, some junior annotators refine the labels based on the prediction results, and each junior annotator is only responsible for part of 100 new data. A coordinator will check the quality of refinement by all junior annotators. For easy cases, the annotation process is over in this stage; for hard cases, senior experts are invited to make more precise annotations. Step \uppercase\expandafter{\romannumeral2} and Step \uppercase\expandafter{\romannumeral3} are repeated until we finish the annotation of all images in our dataset. Finally, we conduct another round of scrutiny for outliers and mistakes and make necessary corrections to ensure the final quality of our dataset. 
In Fig.~\ref{fig-pic-10}, `Junior annotators' are graduate students in the field of medical image analysis. The `Coordinator' is a medical image analysis practitioner with many years of experience, and the `Senior experts' are cooperating doctors in the partner hospital, one of the best orthopedic hospitals in our country.

In total, we have annotations for $1,109$ metal-free CTs and 14 metal-affected CTs. The remaining 61 metal-affected CTs of image are left unannotated and planned for use in unsupervised learning.

\section{Segmentation Methodology}
\label{sect-2}

\begin{figure}[t]
\centering
\includegraphics[width=0.9\textwidth]{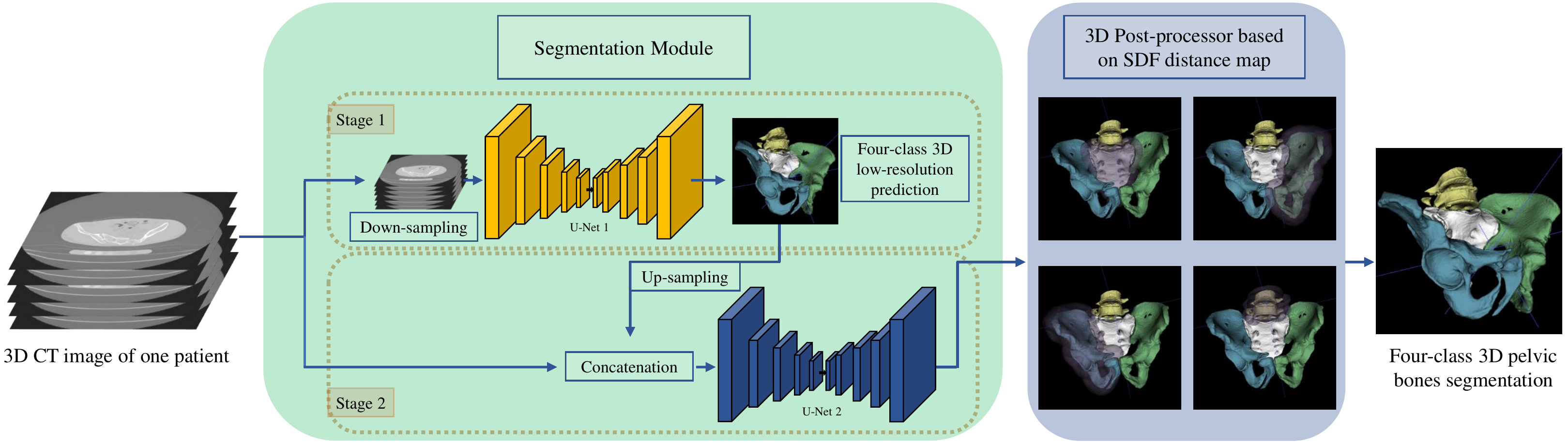}
\caption{Overview of our pelvic bones segmentation system, which learns from multi-domain CT images for effective and robust representations. The 3D U-Net cascade is used here to exploit more spatial information in 3D CT images. SDF is introduced to our post-processor to add distance constraint besides size constraint used in traditional MCR-based method.} \label{fig-pic-5}
\vspace{-0.0cm}
\end{figure}

The overall pipeline of our deep approach for segmenting pelvic bones is illustrated in Fig.~\ref{fig-pic-5}. The input is a 3D CT volume. (i) First, the input is sent to our segmentation module. It is a plug and play (PnP) module that can be replaced at will. 
(ii) After segmentation is done, we send the multi-class 3D prediction to a SDF post-processor, which removes some false predictions and outputs the final multi-bone segmentation result.

\subsection{Segmentation Module}
\label{subsectionSegModule}

Based on our large-scale dataset collected from multiple sources together with annotations, we use a fully supervised method to train a deep network to learn an effective representation of the pelvic bones.
The deep learning framework we choose here is 3D U-Net cascade version of nnU-Net~\cite{FabianNNUnet_a4}, 
which is a robust state-of-the-art deep learning-based medical image segmentation method.
3D U-Net cascade contains two 3D U-net, where the first one is trained on downsampled images (stage 1 in Fig.~\ref{fig-pic-5}), the second one is trained on full resolution images (stage 2 in Fig.~\ref{fig-pic-5}). 
A 3D network can better exploit the useful 3D spatial information in 3D CT images. Training on downsampled images first can enlarge the size of patches in relation to the image, then also enable the 3D network to learn more contextual information. Training on full resolution images second refines the segmentation results predicted from former U-Net.

\subsection{SDF Post Processor}

Post-processing is useful for a stable system in clinical use, preventing some mispredictions in some complex scenes. In the segmentation task, current segmentation systems usually determine whether to remove the outliers according to the size of the connected region to reduce mispredictions. However, in the pelvic fractures scene, broken bones may also be removed as outliers. To this end, we introduce the SDF~\cite{sdf_url1} filtering as our post-processing module to add a \textit{distance constraint} besides the \textit{size constraint}. We calculate SDF based on the maximum connected region (MCR) of 
the anatomical structure in the prediction result, obtaining a 3D distance map that increases from the bone border to the image boundary to help determining whether `outlier prediction' defined by traditional MCR-based method should be removed. 

\section{Experiments}\label{exp}

\subsection{Implementation Details}

We implement our method based on open source code of nnU-Net\footnote{\url{https://github.com/mic-dkfz/nnunet}}~\cite{FabianNNUnet_a4}. We also used MONAI\footnote{\url{https://monai.io/}} during our algorithm development. Details please refer to the Online Resource 1\textsuperscript{\ref {em1}}. For our metal-free dataset, we randomly select 3/5, 1/5, 1/5 cases in each sub-dataset as the training set, validation set, and testing set, respectively, and keep such a data 
partition unchanged in all-dataset experiments and sub-datasets experiments.

\begin{table}[t]
	\centering \small
	\caption{(a) The DC and HD results for different models tested on `ALL' dataset. (b) Effect of different post-processing methods on `ALL' dataset. `ALL’ refers to the six metal-free sub-datasets. `Average' refers to the mean value of four anatomical structures' DC/HD. 
	`Whole' refers to treating Sacrum, Left hip, Right hip and Lumbar spine as a whole bone. The top three numbers in each part are marked in \textbf{bold}, \textcolor{red}{red} and \textcolor{blue}{blue}.}\label{tab_basic}
	\begin{lrbox}{\tablebox}
		\begin{tabular}{lllllllll}
		\begin{tabular}{lllllllll}
			\hline
			Exp & Test  &  Model   & Whole  & Sacrum & Left hip  & Right hip  & Lumbar spine & Average \\
			\cline{4-9}
			& (Dataset)&(Dataset)&Dice/HD&Dice/HD&Dice/HD&Dice/HD&Dice/HD&Dice/HD\\
			\hline
			(a)& ALL &  $\Phi_{ALL(2.5D)}$ & \textcolor{red}{.988}/\textbf{9.28} 
			& \textcolor{red}{.979}/\textcolor{blue}{9.34} 
			& \textbf{.990}/\textbf{3.58} 
			& \textcolor{red}{.990}/\textcolor{red}{3.44}  
			& \textcolor{blue}{.978}/\textcolor{blue}{8.32} 
			& \textcolor{blue}{.984}/\textcolor{red}{6.17} \\
			& ALL & $\Phi_{ALL(3D)}$ 	   & \textcolor{red}{.988}/\textcolor{blue}{11.38} 
			& \textbf{.984}/\textcolor{red}{8.13} 
			& \textcolor{blue}{.988}/\textcolor{blue}{4.99} 
			& \textcolor{red}{.990}/\textcolor{blue}{4.26} 
			& \textcolor{red}{.982}/\textcolor{red}{7.80} 
			& \textcolor{red}{.986}/\textcolor{blue}{6.30}\\
			& ALL & $\Phi_{ALL(3D\_cascade)}$& \textbf{.989}/\textcolor{red}{10.23} 
			& \textbf{.984}/\textbf{7.24} 
			& \textcolor{red}{.989}/\textcolor{red}{4.24} 
			& \textbf{.991}/\textbf{3.03}  
			& \textbf{.984}/\textbf{7.49} 
			& \textbf{.987}/\textbf{5.50}\\
			\hline
			\hline
			(b)& w/o Post & $\Phi_{ALL}$  & \textcolor{red}{.988}/36.27 
			& \textbf{.984}/\textcolor{blue}{38.36} 
			& \textcolor{red}{.988}/\textcolor{blue}{35.43} 
			& \textbf{.991}/28.70 
			& \textcolor{red}{.983}/11.25         
			& \textbf{.987}/28.43\\
			& MCR      & $\Phi_{ALL}$    & \textcolor{red}{.988}/12.93 
			& \textbf{.984}/\textcolor{red}{7.50} 
			& \textbf{.989}/\textbf{4.24}  
			& \textbf{.991}/3.72
			& .978/10.46        
			& \textcolor{red}{.986}/6.48\\
			& SDF(5)& $\Phi_{ALL}$		& \textcolor{red}{.988}/12.02
			& \textbf{.984}/\textbf{7.24} 
			& \textbf{.989}/\textbf{4.24}  
			& \textbf{.991}/3.51
			& \textcolor{blue}{.980}/\textcolor{blue}{9.54}          
			& \textcolor{red}{.986}/6.13\\
			& SDF(15)& $\Phi_{ALL}$		& \textbf{.989}/\textcolor{red}{10.40} 
			& \textbf{.984}/\textbf{7.24} 
			& \textbf{.989}/\textbf{4.24} 
			& \textbf{.991}/\textcolor{red}{3.35}  
			& \textbf{.984}/\textcolor{red}{7.61}   
			& \textbf{.987}/\textcolor{red}{5.61}\\
			& SDF(35)& $\Phi_{ALL}$		& \textbf{.989}/\textbf{10.23} 
			& \textbf{.984}/\textbf{7.24} 
			& \textbf{.989}/\textbf{4.24}  
			& \textbf{.991}/\textbf{3.03}  
			& \textbf{.984}/\textbf{7.49}          
			& \textbf{.987}/\textbf{5.50}\\
			& SDF(55)& $\Phi_{ALL}$		& \textbf{.989}/\textcolor{blue}{10.78}
			& \textbf{.984}/\textbf{7.24} 
			& \textbf{.989}/\textcolor{red}{4.52}  
			& \textbf{.991}/\textcolor{blue}{3.38}  
			& \textbf{.984}/\textbf{7.49}          
			& \textbf{.987}/\textcolor{blue}{5.66}\\
			\hline
		\end{tabular}
		\end{tabular}
	\end{lrbox}
	\scalebox{0.7}[0.7]{\usebox{\tablebox}}
\end{table}

\subsection{Results and Discussion}\label{results}

\subsubsection{Segmentation Module} 
To prove that learning from our large-scale pelvic bones CT dataset is helpful to improve the robustness of our segmentation system, we conduct a series of experiments in different aspects. 

\textbf{Performance of baseline models.}
Firstly, we test the performance of models of different dimensions on our entire dataset. The Exp (a) in Table~\ref{tab_basic} shows the quantitative results. $\Phi_{ALL}$ denotes a deep network model trained on `ALL' dataset. Following the conventions in most literature, we use Dice coefficient(DC) and Hausdorff distance (HD) as the metrics for quantitative evaluation. All results are tested on our testing set. Same as we discussed in Sect.~\ref{subsectionSegModule}, $\Phi_{ALL(3D\_cascade)}$ shows the best performance, achieving an average DC of 0.987 and HD of 5.50 voxels, which means 3D U-Net cascade can learn the semantic features of pelvic anatomy better then 2D/3D U-Net. As the following experiments are all trained with 3D U-Net cascade, the mark $_{(3D\_cascade)}$ of $\Phi_{ALL(3D\_cascade)}$ is omitted for notational clarity.

\begin{table}[t]
	\centering \small
	\caption{The `Average' DC and HD results for different models tested on different datasets. Please refer to the Online Resource 1\textsuperscript{\ref {em1}} for details. The top three numbers in each part are marked in \textbf{bold}, \textcolor{red}{red} and \textcolor{blue}{blue}.}\label{tab_subdataset}
	\begin{lrbox}{\tablebox}
		\begin{tabular}{llllllll}
			\hline
			Average Dice/HD &  &   & Test & Datasets  &   &    \\
			\cline{2-8}
			Models			&ALL &ABDOMEN&COLONOG&MSD\_T10&KITS19&CERVIX&CLINIC\\
			\hline
			
			$\Phi_{ABDOMEN}$&.604/92.81  
			& \textbf{.979}/5.84 
			&.577/104.04 
			&\textcolor{blue}{.980}/3.74
			&.360/158.02
			&.305/92.07
			&.342/148.12\\
			
			$\Phi_{COLONOG}$& \textcolor{red}{.985}/\textcolor{red}{5.84} 
			& \textcolor{blue}{.975}/3.29
			&\textbf{.989}/\textbf{5.65} 
			&.979/4.41
			&\textcolor{blue}{.982}/8.41
			&.969/5.17
			&.974/9.24 \\
			
			$\Phi_{MSD\_T10}$ & .534/96.14 
			& \textbf{.979}/\textcolor{blue}{2.97}
			&.501/106.86 
			&\textbf{.987}/\textcolor{red}{3.36}
			&.245/170.39
			&.085/111.72
			&.261/151.61 \\
			
			$\Phi_{KITS19}$ &.704/68.29     
			& .255/120.31
			&.746/70.75 
			&.267/121.57
			&\textbf{.986}/\textbf{5.65}
			&\textcolor{red}{.973}/5.14
			& \textcolor{blue}{.977}/9.25      \\
			
			$\Phi_{CERVIX}$ &\textcolor{blue}{.973}/\textcolor{blue}{14.75} 
			& .969/4.30
			& \textcolor{blue}{.974}/18.74
			&.967/6.55
			&.979/\textcolor{blue}{7.78}
			& \textcolor{red}{.973}/\textbf{4.49}
			& .974/10.17      \\
			
			$\Phi_{CLINIC}$ & .692/69.89
			& .275/117.09
			&.728/71.93 
			&.254/126.66
			&\textcolor{red}{.985}/11.16
			& .968/9.69
			& \textbf{.983}/\textbf{7.27}\\
			
			$\Phi_{ALL}$ & \textbf{.987}/\textbf{5.50} 
			& \textbf{.979}/\textcolor{red}{2.88} 
			&\textbf{.989}/\textcolor{red}{5.87} 
			&\textbf{.987}/\textbf{3.11}
			&\textcolor{red}{.985}/\textcolor{red}{5.77}
			& \textcolor{blue}{.972}/\textcolor{blue}{5.01} 
			&\textcolor{red}{.982}/\textcolor{red}{7.42}  \\
			
			$\Phi_{ex\ sub-dataset}$& - 
			& \textcolor{red}{.978}/\textbf{2.77} 
			&\textcolor{red}{.986}/\textcolor{blue}{7.37} 
			&\textcolor{red}{.984}/\textcolor{blue}{3.37}
			&\textcolor{blue}{.982}/8.33
			&\textbf{.975}/\textcolor{red}{4.92}
			&.975/\textcolor{blue}{8.87} \\
			\hline
		\end{tabular}
	\end{lrbox}
	\scalebox{0.7}[0.7]{\usebox{\tablebox}}
\end{table}

\begin{figure}[t]
	\centering
	\subfigure[mean DC in Table~\ref{tab_subdataset}]{
		\begin{minipage}[t]{0.5\linewidth}
			\centering
			\includegraphics[width=2in]{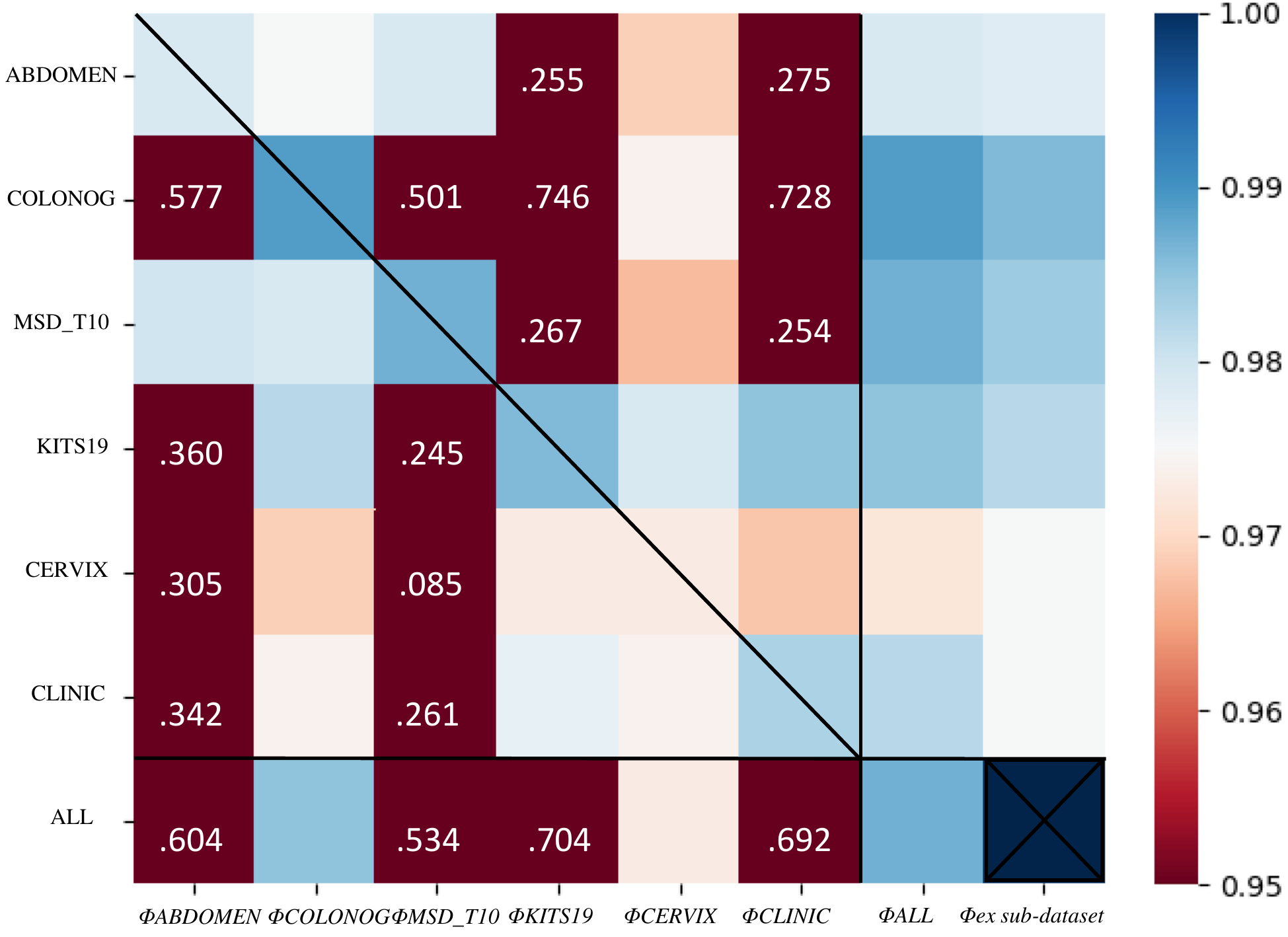}\\
			
		\end{minipage}%
	}%
	\subfigure[mean HD in Table~\ref{tab_subdataset}]{
		\begin{minipage}[t]{0.5\linewidth}
			\centering
			\includegraphics[width=2in]{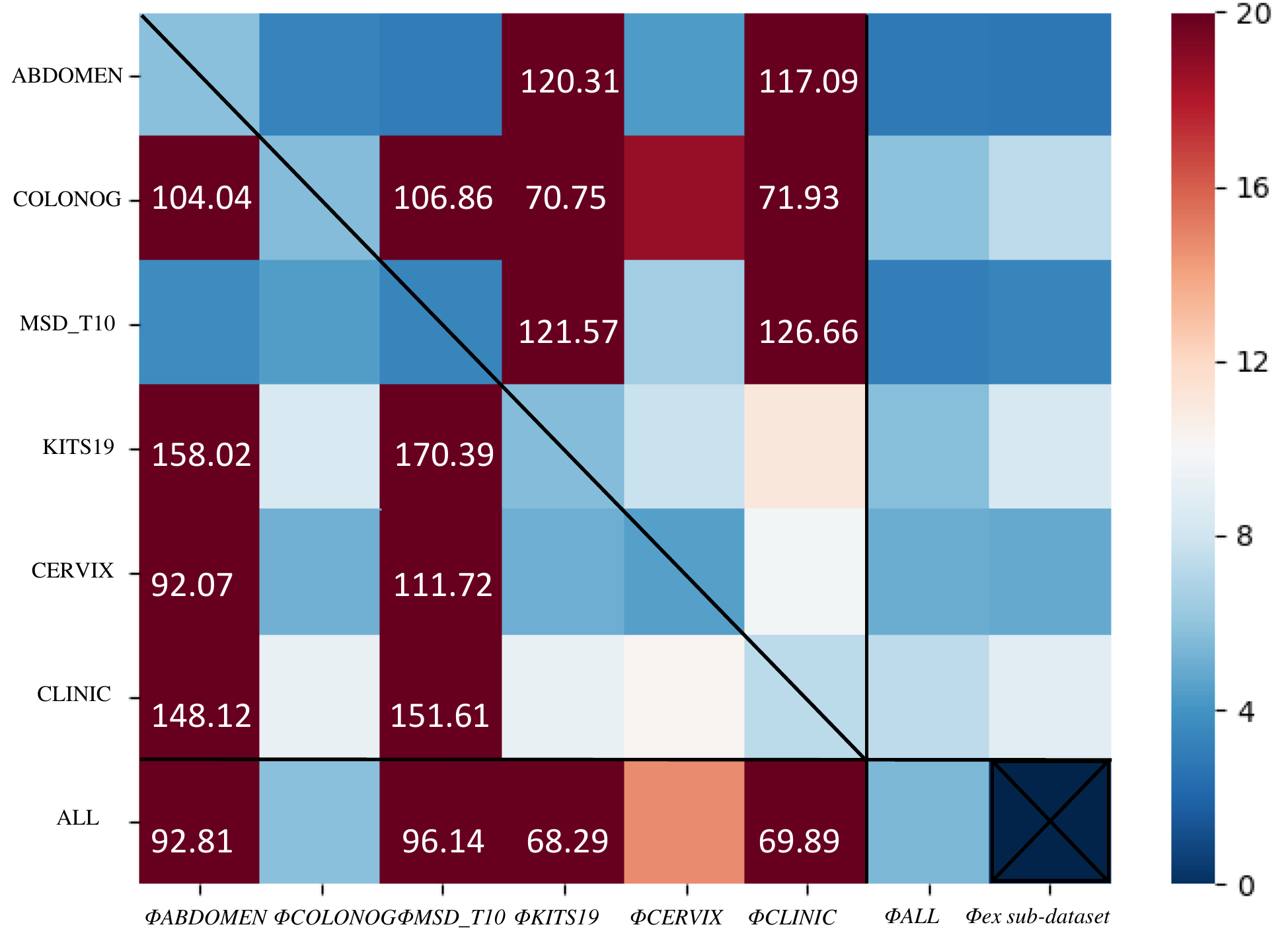}\\
		\end{minipage}%
	}%
	\centering
	\caption{Heat map of DC \& HD results in Table~\ref{tab_subdataset}. The vertical axis represents different sub-datasets and the horizontal axis represents different models. In order to show the normal values more clearly, we clip some outliers to the boundary value, i.e., 0.95 in DC and 30 in HD. The values out of range are marked in the grid. The cross in the lower right corner indicates that there is no corresponding experiment.}
	\vspace{-0.2cm}
	\label{table2fig}
\end{figure}

\begin{figure}[t]
	\includegraphics[width=\textwidth]{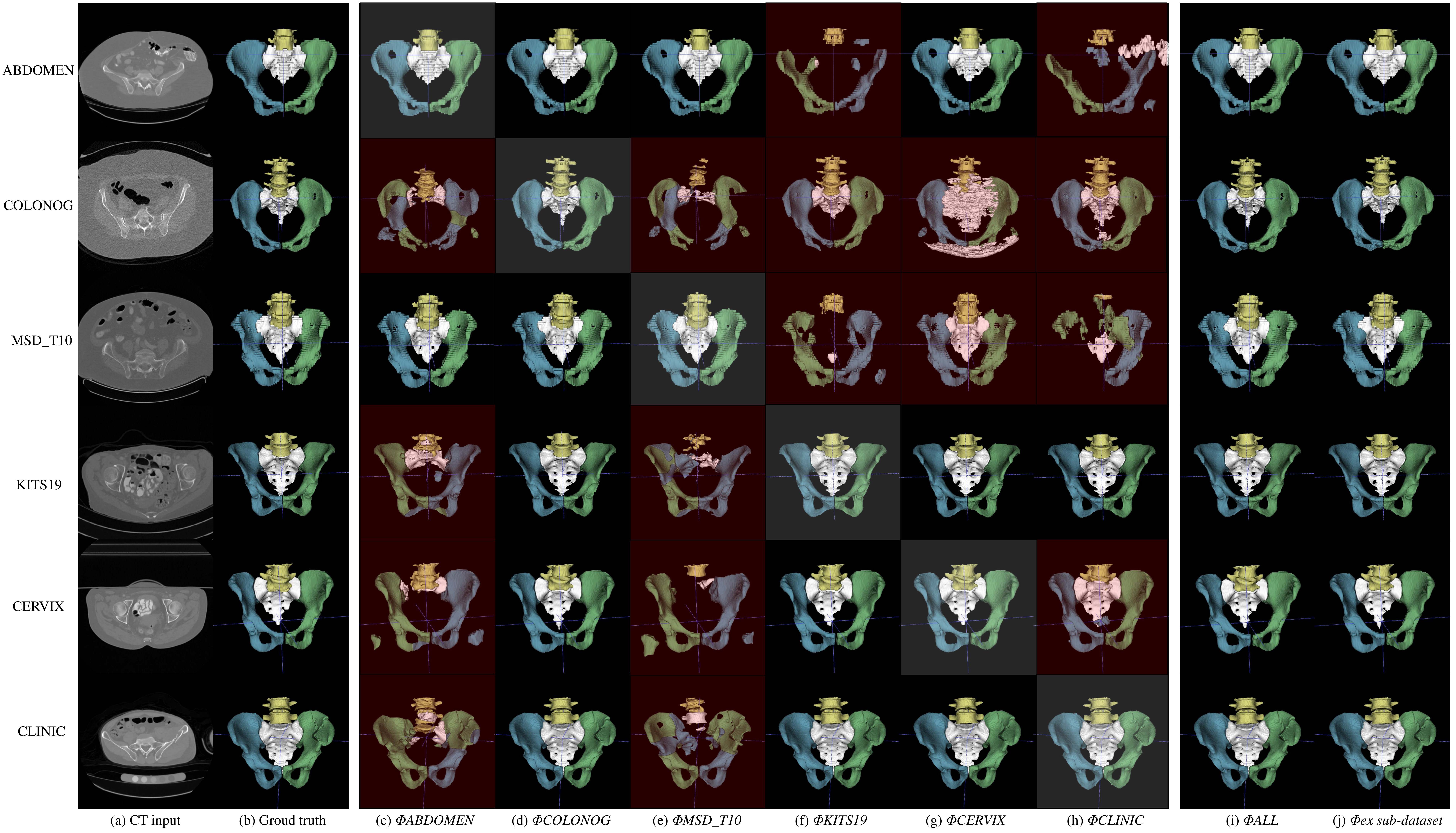}
	\caption{Visualization of segmentation results from six datasets tested on different models. 
		Among them, the white, green, blue and yellow parts of the segmentation results represent the sacrum, left hip bone, right hip bone and lumbar spine, respectively.
	} \label{fig-pic-6}
\end{figure}

\textbf{Generalization across sub-datasets.} Secondly, we train six deep networks, one network per single sub-dataset ($\Phi_{ABDOMEN}$, etc.). Then we test them on each sub-dataset. 
Quantitative and qualitative results are shown in Table~\ref{tab_subdataset} and Fig.~\ref{fig-pic-6}, respectively. We also calculate the performance of $\Phi_{ALL}$ on each sub-dataset. For a fair comparison, cross-testing of sub-dataset networks is also conducted on each sub-dataset's testing set. We observe that the evaluation metrics of model $\Phi_{ALL}$ are generally better than those for the model trained on a single sub-dataset. These models trained on a single sub-dataset are difficult to consistently perform well in other domains, except $\Phi_{COLONOG}$, which contains the largest amount of data from various sources, originally. This observation implies that the domain gap problem does exist and the solution of collecting data directly from multi-source is effective. More intuitively, we show the `Average' values in heat map format in Fig.~\ref{table2fig}. 

Furthermore, we implement \textit{leave-one-out} cross-validation of these six metal-free sub-datasets to verify the generalization ability of this solution. Models are marked as $\Phi_{ex\ ABDOMEN}$, etc. The results of $\Phi_{ex\ COLONOG}$ can fully explain that training with data from multi-sources can achieve good results on data that has not been seen before. When the models trained separately on the other five sub-datasets cannot achieve good results on COLONOG, aggregating these five sub-datasets can get a comparable result compared with $\Phi_{ALL}$, using only one third of the amount of data. More data from multi-sources can be seen as additional constraints on model learning, prompting the network to learn better feature representations of the pelvic bones and the background. In Fig.~\ref{fig-pic-6}, the above discussions can be seen intuitively through qualitative results. 

\textbf{Others.} For more experimental results and discussions, e.g. `Generalization across manufacturers', `Limitations of the dataset', please refer to Online Resource 1\textsuperscript{\ref {em1}}. 

\subsubsection{SDF post-processor}\label{sdfpostprocessor}

\begin{figure}[t]
\includegraphics[width=\textwidth]{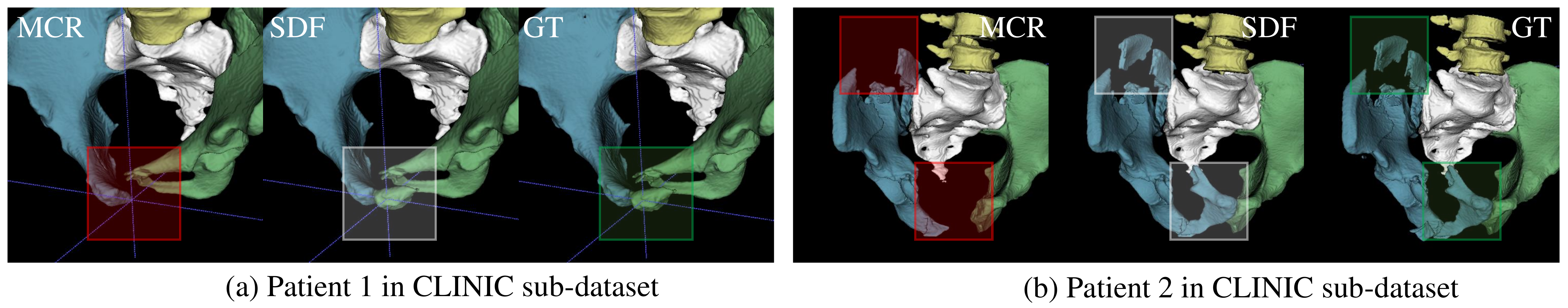}
\caption{Comparison between post-processing methods: traditional MCR and the proposed SDF filtering.} \label{fig-pic-8}
\end{figure}

The Exp (b) in Table~\ref{tab_basic} shows the effect of the post-processing module. SDF post-processor yields a decrease of 80.7\% and 15.1\% in HD compared with no post-processor and MCR post-processor. Details please refer to Online Resource 1\textsuperscript{\ref {em1}}.
The visual effects of two cases are displayed in Fig.~\ref{fig-pic-8}. Large fragments near the anatomical structure are kept with SDF post-processing but are removed by the MCR method.

\section{Conclusion}

To benefit the pelvic surgery and diagnosis community, we curate and open source\textsuperscript{\ref{opensource}} a large-scale pelvic CT dataset pooled from multiple domains, including $1,184$ CT volumes (over 320K CT slices) of various appearance variations, and present a pelvic segmentation system based on deep learning, which, to the best of our knowledge, marks the first attempt in the literature. We train a multi-class network for segmentation of lumbar spine, sacrum, left hip, and right hip using the multiple-domain images to obtain more effective and robust features. SDF filtering further improves the robustness of the system. This system lays a solid foundation for our future work. We plan to test the significance of our system in real clinical practices, and explore more options based on our dataset, e.g. devising a module for metal-affected CTs and domain-independent pelvic bones segmentation algorithm.

%
%

\section*{Declarations}

\paragraph{Funding}
This research was supported in part by the Youth Innovation Promotion Association CAS (grant 2018135).

\paragraph{Conflict of interest}
The authors have no relevant financial or non-financial interests to disclose.

\paragraph{Availability of data and material} Please refer to URL\textsuperscript{\ref{opensource}}.

\paragraph{Code availability} Please refer to URL\textsuperscript{\ref{opensource}}.

\paragraph{Ethical approval} We have obtained the approval from the Ethics Committee of clinical hospital.

\paragraph{Informed consent} Not applicable.


\end{document}